# Estimating Uncertain Spatial Relationships in Robotics*

**Randall Smith**† **Matthew Self**‡ **Peter Cheeseman**§

**SRI International**
**333 Ravenswood Avenue**
**Menlo Park, California 94025**

In this paper, we describe a representation for spatial information, called the *stochastic map*, and associated procedures for building it, reading information from it, and revising it incrementally as new information is obtained. The map contains the estimates of relationships among objects in the map, and their uncertainties, given all the available information. The procedures provide a general solution to the problem of estimating uncertain relative spatial relationships. The estimates are probabilistic in nature, an advance over the previous, very conservative, worst-case approaches to the problem. Finally, the procedures are developed in the context of state-estimation and filtering theory, which provides a solid basis for numerous extensions.

## 1 Introduction

In many applications of robotics, such as industrial automation, and autonomous mobility, there is a need to represent and reason about spatial uncertainty. In the past, this need has been circumvented by special purpose methods such as precision engineering, very accurate sensors and the use of fixtures and calibration points. While these methods sometimes supply sufficient accuracy to avoid the need to represent uncertainty explicitly, they are usually costly. An alternative approach is to use multiple, overlapping, lower resolution sensors and to combine the spatial information (including the uncertainty) from all sources to obtain the best spatial estimate. This integrated information can often supply sufficient accuracy to avoid the need for the hard engineered approach.

In addition to lower hardware cost, the explicit estimation of uncertain spatial information makes it possible to decide *in advance* whether proposed operations are likely to fail because of accumulated uncertainty, and whether proposed sensor information will be sufficient to reduce the uncertainty to tolerable limits. In situations utilizing inexpensive mobile robots,

*The research reported in this paper was supported by the National Science Foundation under Grant ECS-8200615, the Air Force Office of Scientific Research under Contract F49620-84-K-0007, and by General Motors Research Laboratories.

†Currently at General Motors Research Laboratories
Warren, Michigan.

‡Currently at UC Berkeley.

§Currently at NASA Ames Research Center,
Moffett Field, California.

perhaps the *only* way to obtain sufficient accuracy is to combine the (uncertain) information from many sensors. However, a difficulty in combining uncertain spatial information arises because it often occurs in the form of uncertain *relative* information. This is particularly true where many different frames of reference are used, and the uncertain spatial information must be propagated between these frames. This paper presents a general solution to the problem of estimating uncertain spatial relationships, regardless of which frame the information is presented in, or in which frame the answer is required.

Previous methods for representing spatial uncertainty in typical robotic applications (e.g. [Taylor, 1976]) numerically computed min-max bounds on the errors. Brooks developed other methods for computing min-max bounds symbolically[Brooks, 1982]. These min-max approachs are very conservative compared to the probabilistic approach in this paper, because they combine many pieces of information, each with worst case bounds on the errors. More recently, a probabilistic representation of uncertainty was utilized by the HILARE mobile robot [Chatila, 1985] that is similar to the one presented here, except that it uses only a scalar representation of positional uncertainty instead of a multivariate representation of position and orientation. Smith and Cheeseman ([Smith, 1984], [Smith, 1985]), working on problems in off-line programming of industrial automation tasks, proposed operations that could reduce graphs of uncertain relationships (represented by multivariate probability distributions) to a single, best estimate of some relationship of interest. The current paper extends that work, but in the formal setting of estimation theory, and does not utilize graph transformations.

In summary, many important applications require a representation of spatial uncertainty. In addition, methods for combining uncertain spatial information and transforming such information from one frame to another are required. This paper presents a representation that makes explicit the uncertainty of each degree of freedom in the spatial relationships of interest. A method is given for combining uncertain information regardless of which frame it is presented in, and it allows the description of the spatial uncertainty of one frame relative to any other frame. The necessary procedures are presented in matrix form, suitable for efficient implementation. In particular, methods are given for incrementally building the best estimate "map" and its uncertainty as new pieces of uncertain spatial information are added.

# 2 The Stochastic Map

Our knowledge of the spatial relationships among objects is inherently uncertain. A man-made object does not match its geometric model *exactly* because of manufacturing tolerances. *Even if it did*, a sensor could not measure the geometric features, and thus locate the object *exactly*, because of measurement errors. And *even if it could*, a robot using the sensor cannot manipulate the object *exactly* as intended, because of hand positioning errors. These errors can be reduced to neglible limits for some tasks, by "pre-enginerring" the solution — structuring the working environment and using specially–suited high–precision equipment — but at great cost of time and expense. However, rather than treat spatial uncertainty as a side issue in geometrical reasoning, we believe it must be treated as an intrinsic part of spatial representations. In this paper, uncertain spatial relationships will be tied together in a representation called the *stochastic map*. It contains estimates of the spatial relationships, their uncertainties, and their inter-dependencies.

First, the map structure will be described, followed by methods for extracting information from it. Finally, a procedure will be given for building the map *incrementally*, as new spatial information is obtained. To illustrate the theory, we will present an example of a mobile robot acquiring knowledge about its location and the organization of its environment by making sensor observations at different times and in different places.

## 2.1 Representation

In order to formalize the above ideas, we will define the following terms. A *spatial relationship* will be represented by the vector of its *spatial variables*, $\mathbf{x}$. For example, the position and orientation of a mobile robot can be described by its coordinates, $x$ and $y$, in a two dimensional cartesian reference frame and by its orientation, $\phi$, given as a rotation about the $z$ axis:

$$\mathbf{x} = \begin{bmatrix} x \\ y \\ \phi \end{bmatrix}.$$

An *uncertain* spatial relationship, moreover, can be represented by a *probability distribution* over its spatial variables — i.e., by a probability density function that assigns a probability to each particular combination of the spatial variables, $\mathbf{x}$:

$$P(\mathbf{x}) = f(\mathbf{x})d\mathbf{x}.$$

Such detailed knowledge of the probability distribution is usually unneccesary for making decisions, such as whether the robot will be able to complete a given task (e.g. passing through a doorway). Furthermore, most measuring devices provide only a nominal value of the measured relationship, and we can estimate the average error from the sensor specifications. For these reasons, we choose to model an uncertain spatial relationship by estimating the first two moments of its probability distribution—the *mean*, $\hat{\mathbf{x}}$ and the *covariance*, $\mathbf{C}(\mathbf{x})$, defined as:

$$\begin{aligned} \hat{\mathbf{x}} &\triangleq E(\mathbf{x}), \\ \tilde{\mathbf{x}} &\triangleq \mathbf{x} - \hat{\mathbf{x}}, \\ \mathbf{C}(\mathbf{x}) &\triangleq E(\tilde{\mathbf{x}}\tilde{\mathbf{x}}^T). \end{aligned} \tag{1}$$

where $E$ is the expectation operator, and $\tilde{\mathbf{x}}$ is the deviation from the mean. For our mobile robot example, these are:

$$\hat{\mathbf{x}} = \begin{bmatrix} \hat{x} \\ \hat{y} \\ \hat{\phi} \end{bmatrix}, \qquad \mathbf{C}(\mathbf{x}) = \begin{bmatrix} \sigma_x^2 & \sigma_{xy} & \sigma_{x\phi} \\ \sigma_{xy} & \sigma_y^2 & \sigma_{y\phi} \\ \sigma_{x\phi} & \sigma_{y\phi} & \sigma_\phi^2 \end{bmatrix}.$$

The diagonal elements of the covariance matrix are just the variances of the spatial variables, while the off-diagonal elements are the covariances between the spatial variables. It is useful to think of the covariances in terms of their correlation coefficients, $\rho_{ij}$:

$$\rho_{ij} \triangleq \frac{\sigma_{ij}}{\sigma_i \sigma_j} = \frac{E(\tilde{x}_i \tilde{x}_j)}{\sqrt{E(\tilde{x}_i^2)E(\tilde{x}_j^2)}}, \qquad -1 \leq \rho_{ij} \leq 1.$$

Similarly, to model a system of $n$ uncertain spatial relationships, we construct the vector of *all* the spatial variables, which we call the *system state vector*. As before, we will estimate the mean of the state vector, $\hat{\mathbf{x}}$, and the *system covariance matrix*, $\mathbf{C}(\mathbf{x})$:

$$\mathbf{x} = \begin{bmatrix} \mathbf{x}_1 \\ \mathbf{x}_2 \\ \vdots \\ \mathbf{x}_n \end{bmatrix}, \quad \hat{\mathbf{x}} = \begin{bmatrix} \hat{\mathbf{x}}_1 \\ \hat{\mathbf{x}}_2 \\ \vdots \\ \hat{\mathbf{x}}_n \end{bmatrix},$$

$$\mathbf{C}(\mathbf{x}) = \begin{bmatrix} \mathbf{C}(\mathbf{x}_1) & \mathbf{C}(\mathbf{x}_1, \mathbf{x}_2) & \cdots & \mathbf{C}(\mathbf{x}_1, \mathbf{x}_n) \\ \mathbf{C}(\mathbf{x}_2, \mathbf{x}_1) & \mathbf{C}(\mathbf{x}_2) & \cdots & \mathbf{C}(\mathbf{x}_2, \mathbf{x}_n) \\ \vdots & \vdots & \ddots & \vdots \\ \mathbf{C}(\mathbf{x}_n, \mathbf{x}_1) & \mathbf{C}(\mathbf{x}_n, \mathbf{x}_2) & \cdots & \mathbf{C}(\mathbf{x}_n) \end{bmatrix} \tag{2}$$

where:

$$\begin{aligned} \mathbf{C}(\mathbf{x}_i, \mathbf{x}_j) &\triangleq E(\tilde{\mathbf{x}}_i \tilde{\mathbf{x}}_j^T), \\ \mathbf{C}(\mathbf{x}_j, \mathbf{x}_i) &= \mathbf{C}(\mathbf{x}_i, \mathbf{x}_j)^T. \end{aligned} \tag{3}$$

Here, the $\mathbf{x}_i$'s are the vectors of the spatial variables of the individual uncertain spatial relationships, and the $\mathbf{C}(\mathbf{x}_i)$'s are the associated covariance matrices, as discussed earlier. The $\mathbf{C}(\mathbf{x}_i, \mathbf{x}_j)$'s are the cross-covariance matrices between the uncertain spatial relationships. These off–diagonal sub–matrices encode the dependencies between the estimates of the different spatial relationships, and provide the mechanism for updating all the relational estimates that depend on those that are changed.

In our example, each uncertain spatial relationship is of the same form, so $\mathbf{x}$ has $m = 3n$ elements, and we may write:

$$\mathbf{x}_i = \begin{bmatrix} x_i \\ y_i \\ \phi_i \end{bmatrix}, \qquad \hat{\mathbf{x}}_i = \begin{bmatrix} \hat{x}_i \\ \hat{y}_i \\ \hat{\phi}_i \end{bmatrix}, \qquad \mathbf{C}(\mathbf{x}_i, \mathbf{x}_j) = \begin{bmatrix} \sigma_{x_i x_j} & \sigma_{x_i y_j} & \sigma_{x_i \phi_j} \\ \sigma_{x_i y_j} & \sigma_{y_i y_j} & \sigma_{y_i \phi_j} \\ \sigma_{x_i \phi_j} & \sigma_{y_i \phi_j} & \sigma_{\phi_i \phi_j} \end{bmatrix}.$$

Thus our "map" consists of the current estimate of the mean of the system state vector, which gives the nominal locations of objects in the map with respect to the world reference frame, and the associated system covariance matrix, which gives the uncertainty of each point in the map and the inter-dependencies of these uncertainties.

## 2.2 Interpretation

For some decisions based on uncertain spatial relationships, we must assume a particular distribution that fits the estimated moments. For example, a robot might need to be able to calculate the probability that a certain object will be in its field of view, or the probability that it will succeed in passing through a doorway.

Given only the mean, $\mathbf{x}$, and covariance matrix, $\mathbf{C}(\mathbf{x})$, of a multivariate probability distribution, the principle of maximum entropy indicates that the distribution which assumes the least information is the normal distribution. Furthermore if the spatial relationship is calculated by combining many different pieces of information the central limit theorem indicates that the resulting distribution will tend to a normal distribution:

$$P(\mathbf{x}) = \frac{\exp\left[-\frac{1}{2}(\mathbf{x}-\hat{\mathbf{x}})^T \mathbf{C}^{-1}(\mathbf{x})(\mathbf{x}-\hat{\mathbf{x}})\right]}{\sqrt{(2\pi)^m |\mathbf{C}(\mathbf{x})|}} d\mathbf{x}. \tag{4}$$

We will graph uncertain spatial relationships by plotting contours of constant probability from a normal distribution with the given mean and covariance information. These contours are concentric ellipsoids (ellipses for two dimensions) whose parameters can be calculated from the covariance matrix, $\mathbf{C}(\mathbf{x}_i)$ [Nahi, 1976]. It is important to emphasize that we do not assume that the uncertain spatial relationships are described by normal distributions. We estimate the mean and variance of their distributions, and use the normal distribution only when we need to calculate specific probability contours.

In the figures in this paper, the plotted points show the *actual* locations of objects, which are known only by the simulator and displayed for our benefit. The robot's information is shown by the ellipses which are drawn centered on the estimated mean of the relationship and such that they enclose a 99.9% confidence region (about four standard deviations) for the relationships.

## 2.3 Example

Throughout this paper we will refer to a two dimensional example involving the navigation of a mobile robot with three degrees of freedom. In this example the robot performs the following sequence of actions:

- The robot senses object #1
- The robot moves.
- The robot senses an object (object #2) which it determines cannot be object #1.
- Trying again, the robot succeeds in sensing object #1, thus helping to localize itself, object #1, and object #2.

Figure 1 shows two examples of uncertain spatial relationships — the sensed location of object #1, and the end-point of a planned motion for the robot. The robot is initially sitting at a landmark which will be used as the world reference location. There is enough information in our stochastic map at this point for the robot to be able to decide how likely a collision with the object is, if the motion is made. In this case the probability is vanishingly small. The same figure shows how this spatial knowledge can be presented from the robot's new reference frame after its motion. As expected, the uncertainty in the location of object #1 becomes larger when it is compounded with the uncertainty in the robot's motion.

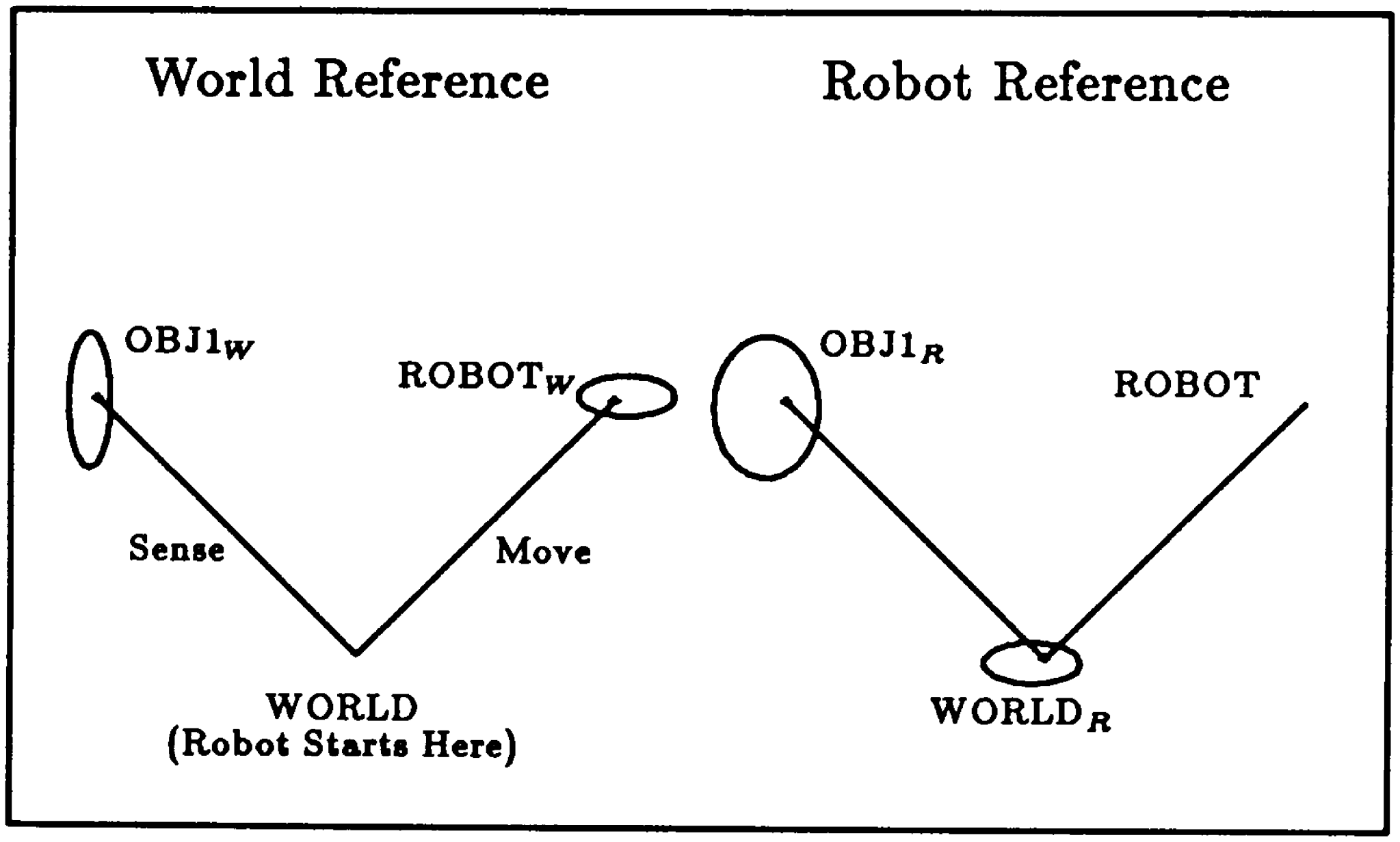

**Figure 1: The Robot Senses Object 1 and Moves**

From this new location, the robot senses object #2 (Figure 2). The robot is able to determine with the information in its stochastic map that this must be a new object and is not object #1 which it observed earlier.

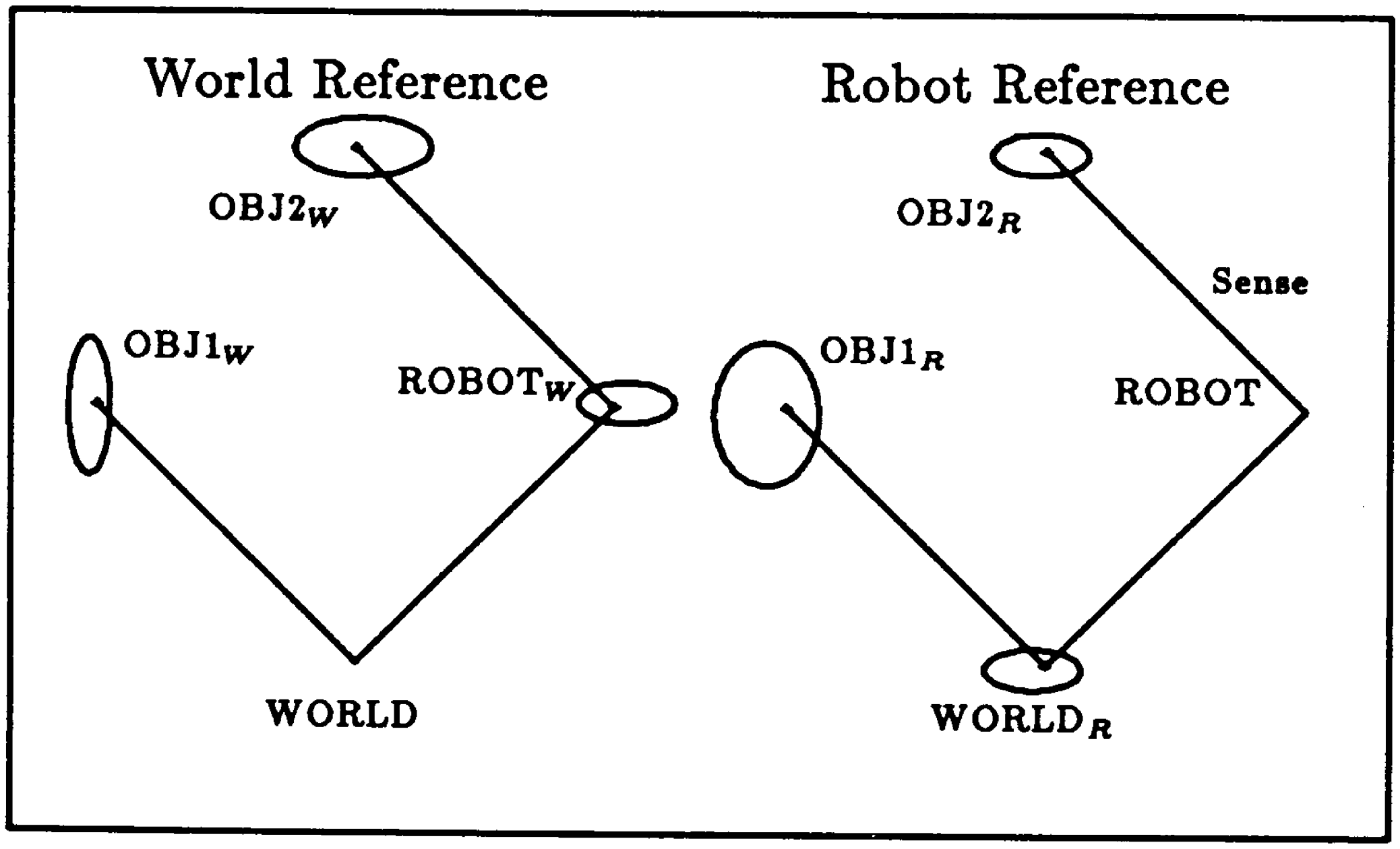

**Figure 2: The Robot Senses Object 2**

In Figure 3, the robot senses object #1 again. This new sensor measurement acts as a constraint and is incorporated into the map, reducing the uncertainty in the locations of the robot, object #1 *and* 0bject #2 (Figure 4).

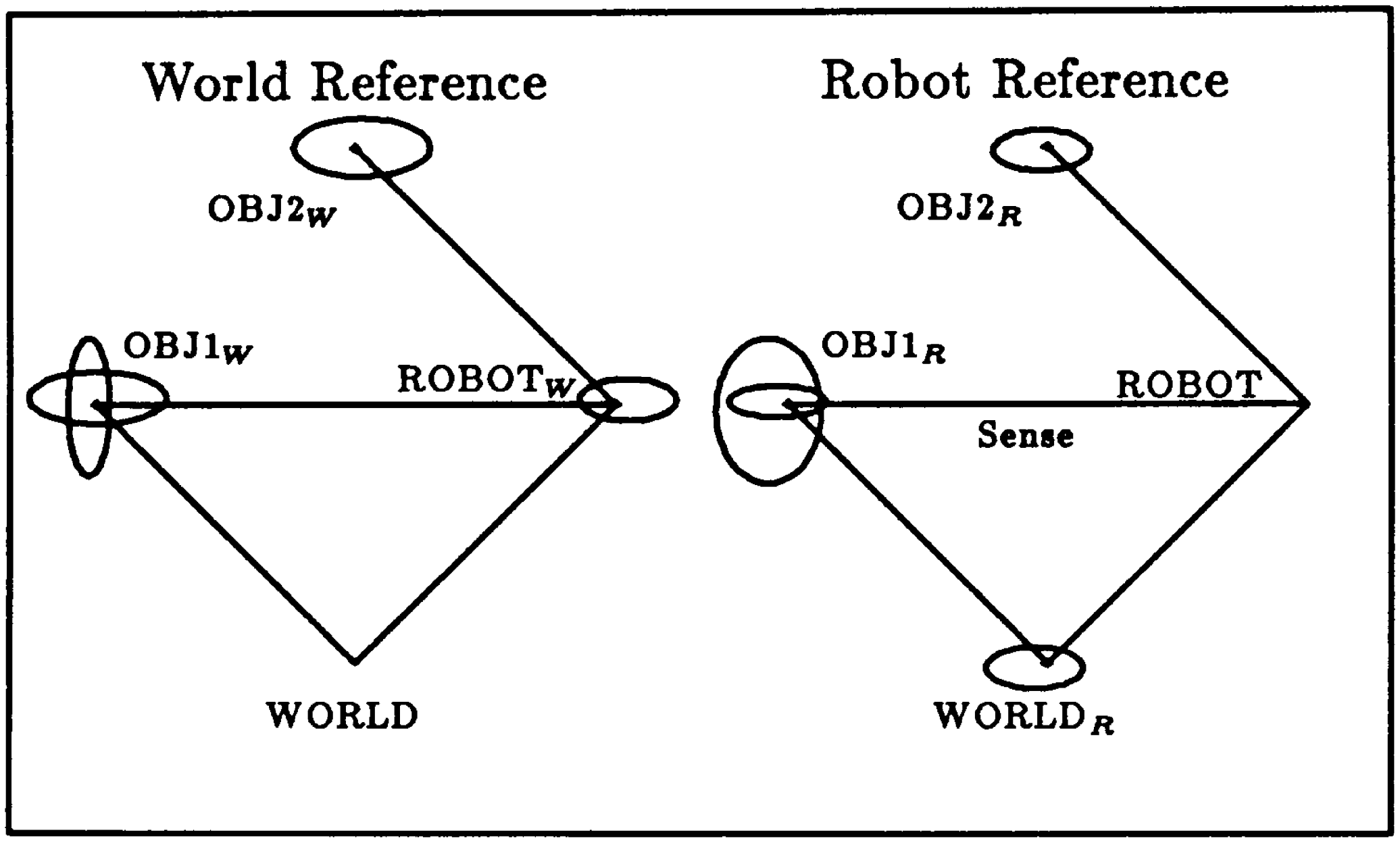

Figure 3: The Robot Senses Object 1 Again

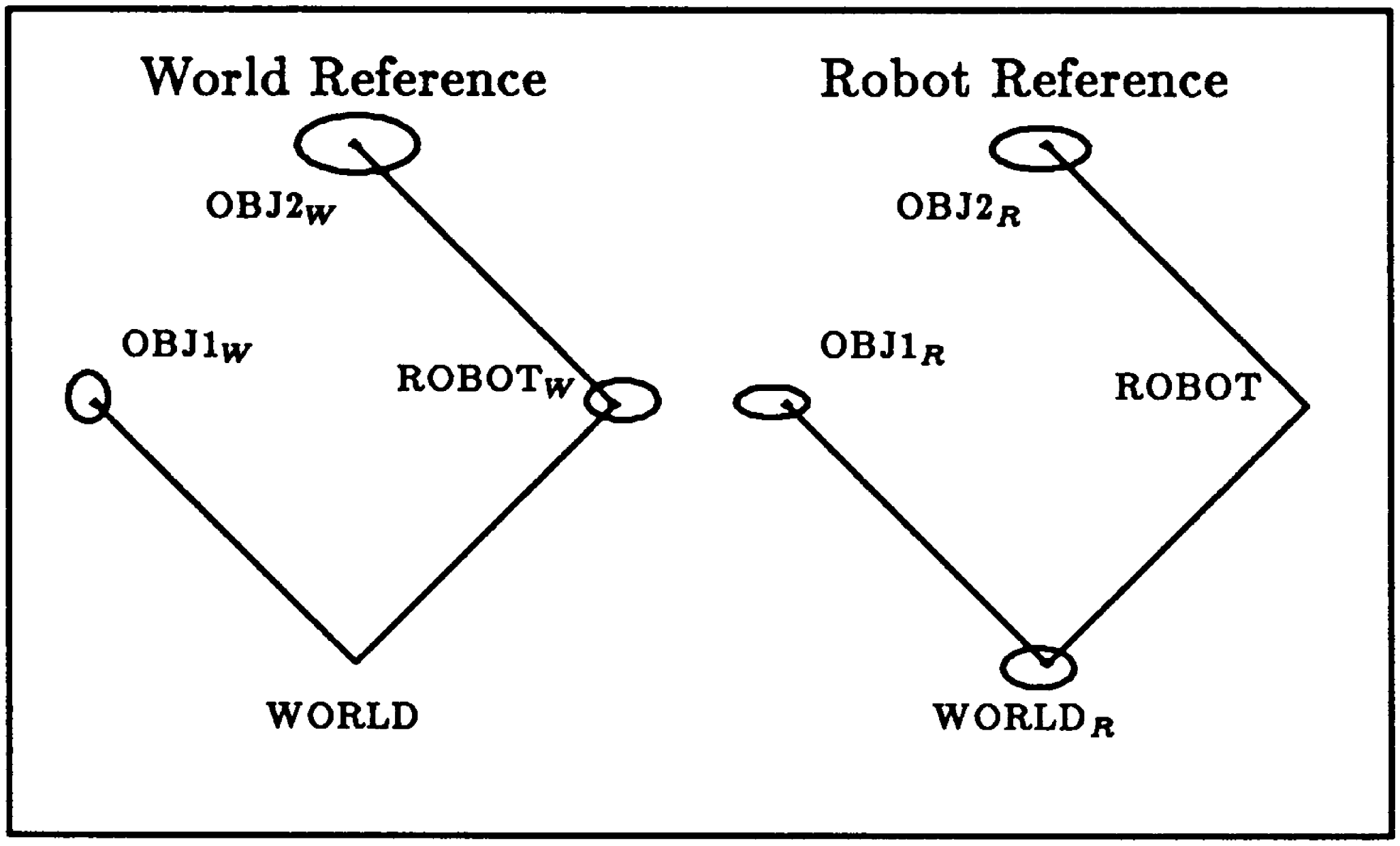

Figure 4: The Updated Estimates After Constraint

# 3 Reading the Map

## 3.1 Uncertain Relationships

Having seen how we can represent uncertain spatial relationships by estimates of the mean and covariance of the system state vector, we now discuss methods for estimating the first two moments of unknown multivariate probability distributions. See [Papoulis, 1965] for detailed justifications of the following topics.

### 3.1.1 Linear Relationships

The simplest case concerns relationships which are linear in the random varables, e.g.:

$$\mathbf{y} = \mathbf{M}\mathbf{x} + \mathbf{b},$$

where, $\mathbf{x}$ $(n \times 1)$ is a random vector, $\mathbf{M}$ $(r \times n)$ is the *non*-random coefficient matrix, $\mathbf{b}$ $(r \times 1)$ is a constant vector, and $\mathbf{y}$ $(r \times 1)$ is the resultant random vector. Using the definitions from (1), and the linearity of the expectation operator, $E$, one can easily verify that the mean of the relationship, $\hat{\mathbf{y}}$, is given by:

$$\hat{\mathbf{y}} = \mathbf{M}\hat{\mathbf{x}} + \mathbf{b}, \tag{5}$$

and the covariance matrix, $\mathbf{C}(\mathbf{y})$, is:

$$\mathbf{C}(\mathbf{y}) = \mathbf{M}\mathbf{C}(\mathbf{x})\mathbf{M}^T. \tag{6}$$

We will also need to be able to compute the covariance between $\mathbf{y}$ and some other relationship, $\mathbf{z}$, given the covariance between $\mathbf{x}$ and $\mathbf{z}$:

$$\begin{aligned} \mathbf{C}(\mathbf{y},\mathbf{z}) &= \mathbf{M}\mathbf{C}(\mathbf{x},\mathbf{z}), \\ \mathbf{C}(\mathbf{z},\mathbf{y}) &= \mathbf{C}(\mathbf{z},\mathbf{x})\mathbf{M}^T. \end{aligned} \tag{7}$$

The first two moments of the multivariate distribution of $\mathbf{y}$ are computed exactly, given correct moments for $\mathbf{x}$. Further, if $\mathbf{x}$ follows a normal distribution, then so does $\mathbf{y}$.

### 3.1.2 Non-Linear Relationships

The first two moments computed by the formulae below for non-linear relationships on random variables will be first-order estimates of the true values. To compute the actual values requires knowledge of the *complete* probability density function of the spatial variables, which will not generally be available in our applications. The usual approach is to approximate the non-linear function

$$\mathbf{y} = \mathbf{f}(\mathbf{x})$$

by a Taylor series expansion about the estimated mean, $\hat{\mathbf{x}}$, yielding:

$$\mathbf{y} = \mathbf{f}(\hat{\mathbf{x}}) + \mathbf{F}_{\mathbf{x}}\tilde{\mathbf{x}} + \cdots,$$

where $\mathbf{F_x}$ is the matrix of partials, or Jacobian, of $\mathbf{f}$ evaluated at $\hat{\mathbf{x}}$:

$$\mathbf{F_x} \triangleq \frac{\partial \mathbf{f}(\mathbf{x})}{\partial \mathbf{x}}(\hat{\mathbf{x}}) \triangleq \begin{bmatrix} \frac{\partial f_1}{\partial x_1} & \frac{\partial f_1}{\partial x_2} & \cdots & \frac{\partial f_1}{\partial x_n} \\ \frac{\partial f_2}{\partial x_1} & \frac{\partial f_2}{\partial x_2} & \cdots & \frac{\partial f_2}{\partial x_n} \\ \vdots & \vdots & \ddots & \vdots \\ \frac{\partial f_r}{\partial x_1} & \frac{\partial f_r}{\partial x_2} & \cdots & \frac{\partial f_r}{\partial x_n} \end{bmatrix}_{\mathbf{x}=\hat{\mathbf{x}}}.$$

This terminology is the extension of the $f_x$ terminology from scalar calculus to vectors. The Jacobians are always understood to be evaluated at the estimated mean of the given variables.

Truncating the expansion for $\mathbf{y}$ after the linear term, and taking the expectation produces the linear estimate of the mean of $\mathbf{y}$:

$$\hat{\mathbf{y}} \approx \mathbf{f}(\hat{\mathbf{x}}). \tag{8}$$

Similarly, the first-order estimate of the covariances are:

$$\begin{aligned} \mathbf{C}(\mathbf{y}) &\approx \mathbf{F_x}\mathbf{C}(\mathbf{x})\mathbf{F_x}^T, \\ \mathbf{C}(\mathbf{y},\mathbf{z}) &\approx \mathbf{F_x}\mathbf{C}(\mathbf{x},\mathbf{z}), \\ \mathbf{C}(\mathbf{z},\mathbf{y}) &\approx \mathbf{C}(\mathbf{z},\mathbf{x})\mathbf{F_x}^T. \end{aligned} \tag{9}$$

Though not utilized in our application, the second order term may be included in the Taylor series expansion to improve the mean estimate:

$$\mathbf{y} = \mathbf{f}(\hat{\mathbf{x}}) + \mathbf{F_x}\tilde{\mathbf{x}} + \frac{1}{2}\mathbf{F_{xx}}(\tilde{\mathbf{x}}\tilde{\mathbf{x}}^T) + \cdots,$$

We denote the (3 dimensional) matrix of second partials of $\mathbf{f}$ by $\mathbf{F_{xx}}$. To avoid unecce-sary complexity, we simply state that the $i$th element of the vector produced when $\mathbf{F_{xx}}$ is multiplied on the right by a matrix $\mathbf{A}$ is defined by:

$$(\mathbf{F_{xx}}\mathbf{A})_i = trace\left[\left(\left.\frac{\partial^2 f_i}{\partial x_j \partial x_k}\right|_{\mathbf{x}=\hat{\mathbf{x}}}\right)\mathbf{A}\right].$$

The second-order estimate of the mean of $\mathbf{y}$ is then:

$$\hat{\mathbf{y}} \approx \mathbf{f}(\hat{\mathbf{x}}) + \frac{1}{2}\mathbf{F_{xx}}\mathbf{C}(\mathbf{x}),$$

and the second-order estimate of the covariance is:

$$\mathbf{C}(\mathbf{y}) \approx \mathbf{F_x}\mathbf{C}(\mathbf{x})\mathbf{F_x}^T - \frac{1}{4}\mathbf{F_{xx}}\mathbf{C}(\mathbf{x})\mathbf{C}(\mathbf{x})^T\mathbf{F_{xx}}^T.$$

In the remainder of this paper we consider only first order estimates, and the symbol "$\approx$" should read as "linear estimate of."

## 3.2 Spatial Relationships

We now consider the actual spatial relationships which are most often encountered in robotics applications. We will develop our presentation about the three degree of freedom formulae, since they suit our examples concerning a mobile robot. Formulae for the three dimensional case with six degrees of freedom are given in Appendix A.

We would like to take a chain of relationships, starting at an initial coordinate frame, passing through several intermediate frames to a final frame, and estimate the resultant relationship between initial and final frames. Since frame relationships are directed, we will need the ability to invert the sense of some given relationships during the calculation. The formulae needed for calculating these estimates are given in the following sections.

### 3.2.1 Compounding

Given two spatial relationships, $\mathbf{x}_{ij}$ and $\mathbf{x}_{jk}$, as in Figure 1 (under Robot Reference), we wish to compute the resultant relationship $\mathbf{x}_{ik}$. The formula for computing $\mathbf{x}_{ik}$ from $\mathbf{x}_{ij}$ and $\mathbf{x}_{jk}$ is:

$$\mathbf{x}_{ik} \stackrel{\triangle}{=} \mathbf{x}_{ij} \oplus \mathbf{x}_{jk} = \begin{bmatrix} x_{jk}\cos\phi_{ij} - y_{jk}\sin\phi_{ij} + x_{ij} \\ x_{jk}\sin\phi_{ij} + y_{jk}\cos\phi_{ij} + y_{ij} \\ \phi_{ij} + \phi_{jk} \end{bmatrix}.$$

We call this operation compounding, and it is used to calculate the resultant relationship from two given relationships which are arranged head-to-tail. It would be used, for instance, to determine the location of a mobile robot after a sequence of relative motions. Remember that these transformations involve rotations, so compounding is not merely vector addition.

Utilizing (8), the first-order estimate of the mean of the compounding operation is:

$$\hat{\mathbf{x}}_{ik} \approx \hat{\mathbf{x}}_{ij} \oplus \hat{\mathbf{x}}_{jk}.$$

Also, from (9), the first-order estimate of the covariance is:

$$\mathbf{C}(\mathbf{x}_{ik}) \approx \mathbf{J}_{\oplus} \begin{bmatrix} \mathbf{C}(\mathbf{x}_{ij}) & \mathbf{C}(\mathbf{x}_{ij}, \mathbf{x}_{jk}) \\ \mathbf{C}(\mathbf{x}_{jk}, \mathbf{x}_{ij}) & \mathbf{C}(\mathbf{x}_{jk}) \end{bmatrix} \mathbf{J}_{\oplus}^T.$$

where the Jacobian of the compounding operation, $\mathbf{J}_{\oplus}$ is given by:

$$\mathbf{J}_{\oplus} \stackrel{\triangle}{=} \frac{\partial(\mathbf{x}_{ij} \oplus \mathbf{x}_{jk})}{\partial(\mathbf{x}_{ij}, \mathbf{x}_{jk})} = \frac{\partial \mathbf{x}_{ik}}{\partial(\mathbf{x}_{ij}, \mathbf{x}_{jk})} = \begin{bmatrix} 1 & 0 & -(y_{ik} - y_{ij}) & \cos\phi_{ij} & -\sin\phi_{ij} & 0 \\ 0 & 1 & (x_{ik} - x_{ij}) & \sin\phi_{ij} & \cos\phi_{ij} & 0 \\ 0 & 0 & 1 & 0 & 0 & 1 \end{bmatrix}.$$

Note how we have utilized the resultant relationship $\mathbf{x}_{ik}$ in expressing the Jacobian. This results in greater computational efficiency than expressing the Jacobian only in terms of the compounded relationships $\mathbf{x}_{ij}$ and $\mathbf{x}_{jk}$. We can always estimate the mean of an uncertain relationship and then use this result when evaluating the Jacobian to estimate the covariance of the relationship.

If the two relationships being compounded are independent ($\mathbf{C}(\mathbf{x}_{ij}, \mathbf{x}_{jk}) = \mathbf{0}$), we can rewrite the first-order estimate of the covariance as:

$$\mathbf{C}(\mathbf{x}_{ik}) \approx \mathbf{J}_{1\oplus}\mathbf{C}(\mathbf{x}_{ij})\mathbf{J}_{1\oplus}^T + \mathbf{J}_{2\oplus}\mathbf{C}(\mathbf{x}_{jk})\mathbf{J}_{2\oplus}^T$$

where $\mathbf{J}_{1\oplus}$ and $\mathbf{J}_{2\oplus}$ are the left and right halves ($3 \times 3$) of the compounding Jacobian ($3 \times 6$):

$$\mathbf{J}_{\oplus} = \begin{bmatrix} \mathbf{J}_{1\oplus} & \mathbf{J}_{2\oplus} \end{bmatrix}.$$

### 3.2.2 The Inverse Relationship

Given a relationship $\mathbf{x}_{ij}$, the formula for the coordinates of the inverse relationship $\mathbf{x}_{ji}$, as a function of $\mathbf{x}_{ij}$ is:

$$\mathbf{x}_{ji} \stackrel{\triangle}{=} \ominus\mathbf{x}_{ij} \stackrel{\triangle}{=} \begin{bmatrix} -x_{ij}\cos\phi_{ij} - y_{ij}\sin\phi_{ij} \\ x_{ij}\sin\phi_{ij} - y_{ij}\cos\phi_{ij} \\ -\phi_{ij} \end{bmatrix}.$$

We call this the reverse relationship. Using (8) we get the first-order mean estimate:

$$\hat{\mathbf{x}}_{ji} \approx \ominus\hat{\mathbf{x}}_{ij}.$$

From (9) the first-order covariance estimate is:

$$\mathbf{C}(\mathbf{x}_{ji}) \approx \mathbf{J}_{\ominus}\mathbf{C}(\mathbf{x}_{ij})\mathbf{J}_{\ominus}^{T},$$

where the Jacobian for the reversal operation, $\mathbf{J}_{\ominus}$ is:

$$\mathbf{J}_{\ominus} \stackrel{\triangle}{=} \frac{\partial\mathbf{x}_{ji}}{\partial\mathbf{x}_{ij}} = \begin{bmatrix} -\cos\phi_{ij} & -\sin\phi_{ij} & y_{ji} \\ \sin\phi_{ij} & -\cos\phi_{ij} & -x_{ji} \\ 0 & 0 & -1 \end{bmatrix}.$$

Note that the uncertainty is not inverted, but rather expressed from the opposite (reverse) point of view.

### 3.2.3 Composite Relationships

We have shown how to compute the resultant of two relationships which are arranged head-to-tail, and also how to reverse a relationship. With these two operations we can calculate the resultant of any sequence of relationships. For example, the resultant of a chain of relationships arranged head-to-tail can be computed recursively by:

$$\begin{aligned} \mathbf{x}_{il} &= \mathbf{x}_{ij} \oplus \mathbf{x}_{jl} = \mathbf{x}_{ij} \oplus (\mathbf{x}_{jk} \oplus \mathbf{x}_{kl}) \\ &= \mathbf{x}_{ik} \oplus \mathbf{x}_{kl} = (\mathbf{x}_{ij} \oplus \mathbf{x}_{jk}) \oplus \mathbf{x}_{kl} \end{aligned}$$

Note, the compounding operation is associative, but not commutative. We have denoted the reversal operation by $\ominus$ so that by analogy to conventional $+$ and $-$ we may write:

$$\mathbf{x}_{ij} \ominus \mathbf{x}_{kj} \stackrel{\triangle}{=} \mathbf{x}_{ij} \oplus (\ominus\mathbf{x}_{kj}).$$

This is the head-to-head combination of two relationships. The tail-to-tail combination arises quite often (as in Figure 1, under World Reference), and is given by:

$$\mathbf{x}_{jk} = \ominus\mathbf{x}_{ij} \oplus \mathbf{x}_{ik}$$

To estimate the mean of a complex relationship, such as the tail-to-tail combination, we merely solve the estimate equations recursively:

$$\hat{\mathbf{x}}_{jk} = \hat{\mathbf{x}}_{ji} \oplus \hat{\mathbf{x}}_{ik} = \ominus\hat{\mathbf{x}}_{ij} \oplus \hat{\mathbf{x}}_{ik}.$$

The covariance can be estimated in a similar way:

$$\begin{aligned}
\mathbf{C}(\mathbf{x}_{jk}) &\approx \mathbf{J}_{\oplus} \begin{bmatrix} \mathbf{C}(\mathbf{x}_{ji}) & \mathbf{C}(\mathbf{x}_{ji}, \mathbf{x}_{ik}) \\ \mathbf{C}(\mathbf{x}_{ik}, \mathbf{x}_{ji}) & \mathbf{C}(\mathbf{x}_{ik}) \end{bmatrix} \mathbf{J}_{\oplus}^T \\
&\approx \mathbf{J}_{\oplus} \begin{bmatrix} \mathbf{J}_{\ominus}\mathbf{C}(\mathbf{x}_{ij})\mathbf{J}_{\ominus}^T & \mathbf{J}_{\ominus}\mathbf{C}(\mathbf{x}_{ij}, \mathbf{x}_{ik}) \\ \mathbf{C}(\mathbf{x}_{ik}, \mathbf{x}_{ij})\mathbf{J}_{\ominus}^T & \mathbf{C}(\mathbf{x}_{ik}) \end{bmatrix} \mathbf{J}_{\oplus}^T.
\end{aligned}$$

This method is easy to implement as a recursive algorithm. An equivalent method is to precompute the Jacobians of useful combinations of relationships such as the tail-to-tail combination by using the chain rule. Thus, the Jacobian of the tail-to-tail relationship, ${}_{\ominus}\mathbf{J}_{\oplus}$, is given by:

$${}_{\ominus}\mathbf{J}_{\oplus} \triangleq \frac{\partial \mathbf{x}_{jk}}{\partial(\mathbf{x}_{ij}, \mathbf{x}_{ik})} = \frac{\partial \mathbf{x}_{jk}}{\partial(\mathbf{x}_{ji}, \mathbf{x}_{ik})} \frac{\partial(\mathbf{x}_{ji}, \mathbf{x}_{ik})}{\partial(\mathbf{x}_{ij}, \mathbf{x}_{ik})} = \mathbf{J}_{\oplus} \begin{bmatrix} \mathbf{J}_{\ominus} & \mathbf{0} \\ \mathbf{0} & \mathbf{I} \end{bmatrix} = \begin{bmatrix} \mathbf{J}_{1\oplus}\mathbf{J}_{\ominus} & \mathbf{J}_{2\oplus} \end{bmatrix}.$$

Comparison will show that these two methods are symbolically equivalent, but the recursive method is easier to program, while pre-computing the composite Jacobians is more computationally efficient. Even greater computational efficiency can be achieved by making a change of variables such that the already computed mean estimate is used to evaluate the Jacobian, much as described earlier and in Appendix A.

It may appear that we are calculating first-order estimates of first-order estimates of ..., but actually this recursive procedure produces *precisely* the same result as calculating the first-order estimate of the composite relationship. This is in contrast to min-max methods which make conservative estimates at each step and thus produce *very* conservative estimates of a composite relationship.

If we now assume that the cross-covariance terms in the estimate of the covariance of the tail-to-tail relationship are zero, we get:

$$\mathbf{C}(\mathbf{x}_{jk}) \approx \mathbf{J}_{1\oplus}\mathbf{J}_{\ominus}\mathbf{C}(\mathbf{x}_{ij})\mathbf{J}_{\ominus}^T\mathbf{J}_{1\oplus}^T + \mathbf{J}_{2\oplus}\mathbf{C}(\mathbf{x}_{ik})\mathbf{J}_{2\oplus}^T$$

The Jacobians for six degree-of-freedom compounding and reversal relationships are given in Appendix A.

### 3.2.4 Extracting Relationships

We have now developed enough machinery to describe the procedure for estimating the relationships between objects which are in our map. The map contains, by definition, estimates of the locations of objects with respect to the world frame; these relations can be extracted directly. Other relationships are implicit, and must be extracted, using methods developed in the previous sections. For any general spatial relationship among world locations we can write:

$$\mathbf{y} = \mathbf{g}(\mathbf{x}).$$

The estimated mean and covariance of the relationship are given by:

$$\begin{aligned}
\hat{\mathbf{y}} &\approx \mathbf{g}(\hat{\mathbf{x}}), \\
\mathbf{C}(\mathbf{y}) &\approx \mathbf{G}_{\mathbf{x}}\mathbf{C}(\mathbf{x})\mathbf{G}_{\mathbf{x}}^T.
\end{aligned}$$

In our mobile robot example we will need to be able to estimate the relative location of one object with respect to the coordinate frame of another object in our map. In this case, we would simply substitute the tail-to-tail operation previously discussed for $\mathbf{g}()$,

$$\mathbf{y} = \mathbf{x}_{ij} = \ominus\mathbf{x}_i \oplus \mathbf{x}_j.$$

# 4 Building the Map

Our map represents uncertain spatial relationships among objects referenced to a common world frame. Entries in the map may change for two reasons:

- An object moves.
- New spatial information is obtained.

To change the map, we must change the two components that define it — the (mean) estimate of the system state vector, $\hat{\mathbf{x}}$, and the estimate of the system variance matrix, $\mathbf{C}(\mathbf{x})$. Figure 5 shows the changes in the system due to moving objects, or the addition of new spatial information (from sensing).

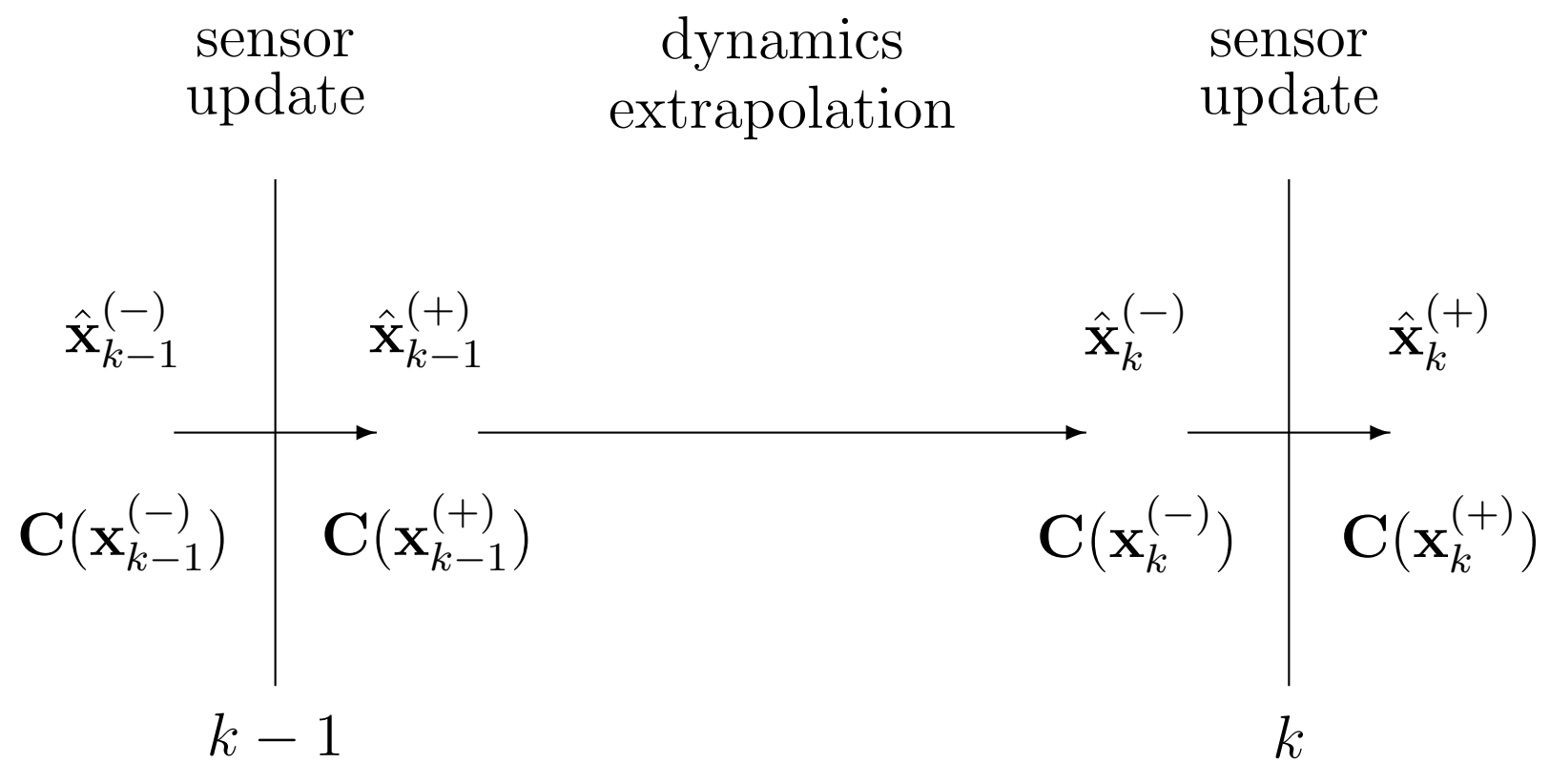

Figure 1: The Changing Map

We will assume that new spatial information is obtained at discrete moments, marked by states $k$. The update of the estimates at state $k$, based on new information, is considered to be instantaneous. The estimates, at state $k$, *prior* to the integration of the new information are denoted by $\hat{\mathbf{x}}_k^{(-)}$ and $\mathbf{C}(\mathbf{x}_k^{(-)})$, and *after* the integration by $\hat{\mathbf{x}}_k^{(+)}$ and $\mathbf{C}(\mathbf{x}_k^{(+)})$.

In the interval between states the system may be changing dynamically — for instance, the robot may be moving. When an object moves, we must define a process to extrapolate the estimate of the state vector and uncertainty at state $k-1$, to state $k$ to reflect the changing relationships.

## 4.1 Moving Objects

Before describing how the map changes as the mobile robot moves, we will present the general case, which treats any processes that change the state of the system.

The *system dynamics model*, or process model, describes how components of the system state vector change (as a function of time in a continuous system, or by discrete transitions). Between state $k-1$ and $k$, no measurements of external objects are made. The new state is determined only by the process model, $\mathbf{f}$, as a function of the old state, and any control variables applied in the process (such as relative motion commands sent to our mobile robot). The process model is thus:

$$\mathbf{x}_k^{(-)} = \mathbf{f}\left(\mathbf{x}_{k-1}^{(+)}, \mathbf{y}_{k-1}\right), \tag{10}$$

where $\mathbf{y}$ is a vector comprised of control variables, $\mathbf{u}$, corrupted by mean-zero process noise, $\mathbf{w}$, with covariance $\mathbf{C}(\mathbf{w})$. That is, $\mathbf{y}$ is a noisy control input to the process, given by:

$$\mathbf{y} = \mathbf{u} + \mathbf{w}. \tag{11}$$

$$\hat{\mathbf{y}} = \mathbf{u}, \qquad \mathbf{C}(\mathbf{y}) = \mathbf{C}(\mathbf{w}).$$

Given the estimates of the state vector and variance matrix at state $k-1$, the estimates are extrapolated to state $k$ by:

$$\hat{\mathbf{x}}_k^{(-)} \approx \mathbf{f}\left(\hat{\mathbf{x}}_{k-1}^{(+)}, \hat{\mathbf{y}}_{k-1}\right), \tag{12}$$

$$\mathbf{C}(\mathbf{x}_k^{(-)}) \approx \mathbf{F}_{(\mathbf{x},\mathbf{y})} \begin{bmatrix} \mathbf{C}(\mathbf{x}_{k-1}^{(+)}) & \mathbf{C}(\mathbf{x}_{k-1}^{(+)}, \mathbf{y}_{k-1}) \\ \mathbf{C}(\mathbf{y}_{k-1}, \mathbf{x}_{k-1}^{(+)}) & \mathbf{C}(\mathbf{y}_{k-1}) \end{bmatrix} \mathbf{F}_{(\mathbf{x},\mathbf{y})}^T.$$

where,

$$\mathbf{F}_{(\mathbf{x},\mathbf{y})} = \begin{bmatrix} \mathbf{F}_{\mathbf{x}} & \mathbf{F}_{\mathbf{y}} \end{bmatrix} \triangleq \frac{\partial \mathbf{f}(\mathbf{x},\mathbf{y})}{\partial(\mathbf{x},\mathbf{y})} \left(\hat{\mathbf{x}}_{k-1}^{(+)}, \hat{\mathbf{y}}_{k-1}\right)$$

If the process noise is uncorrelated with the state, then the off-diagonal sub-matrices in the matrix above are $\mathbf{0}$ and the covariance estimate simplifies to:

$$\mathbf{C}(\mathbf{x}_k^{(-)}) \approx \mathbf{F}_{\mathbf{x}} \mathbf{C}(\mathbf{x}_{k-1}^{(+)}) \mathbf{F}_{\mathbf{x}}^T + \mathbf{F}_{\mathbf{y}} \mathbf{C}(\mathbf{y}_{k-1}) \mathbf{F}_{\mathbf{y}}^T.$$

The new state estimates become the current estimates to be extrapolated to the next state, and so on.

In our example, only the robot moves, so the process model need only describe its motion. A continuous dynamics model can be developed given *a particular robot*, and the above equations can be reformulated as functions of time (see [Gelb, 1984]). However, if the robot only makes sensor observations at discrete times, then the discrete motion approximation is quite adequate. When the robot moves, it changes its relationship, $\mathbf{x}_R$, with the world. The robot makes an uncertain relative motion, $\mathbf{y}_R = \mathbf{u}_R + \mathbf{w}_R$, to reach a final world location $\mathbf{x}'_R$. Thus,

$$\mathbf{x}'_R = \mathbf{x}_R \oplus \mathbf{y}_R.$$

Only a small portion of the map needs to be changed due to the change in the robot's location from state to state — specifically, the $R$th element of the estimated mean of the state vector, and the $R$th row and column of the estimated variance matrix. Thus, $\hat{\mathbf{x}}_{k-1}^{(+)}$ becomes $\hat{\mathbf{x}}_{k}^{(-)}$ and $\mathbf{C}(\mathbf{x}_{k-1}^{(+)})$ becomes $\mathbf{C}(\mathbf{x}_{k}^{(-)})$, as shown below:

$$\hat{\mathbf{x}}_{k-1}^{(+)} = \left[\begin{array}{c} \\ \hline \hat{\mathbf{x}}_R \\ \hline \\ \\ \end{array}\right], \quad \hat{\mathbf{x}}_{k}^{(-)} = \left[\begin{array}{c} \\ \hline \hat{\mathbf{x}}'_R \\ \hline \\ \\ \end{array}\right], \quad \mathbf{C}(\mathbf{x}_{k}^{(-)}) = \left[\begin{array}{c|c|c} & \mathbf{B}'^T & \\ \hline \mathbf{B}' & \mathbf{A}' & \\ \hline & & \\ & & \end{array}\right]$$

where

$$\hat{\mathbf{x}}'_R \approx \hat{\mathbf{x}}_R \oplus \hat{\mathbf{y}}_R,$$
$$\mathbf{A}' = \mathbf{C}(\mathbf{x}'_R) \approx \mathbf{J}_{1\oplus}\mathbf{C}(\mathbf{x}_R)\mathbf{J}_{1\oplus}^T + \mathbf{J}_{2\oplus}\mathbf{C}(\mathbf{y}_R)\mathbf{J}_{2\oplus}^T,$$
$$\mathbf{B}'_i = \mathbf{C}(\mathbf{x}'_R, \mathbf{x}_i) \approx \mathbf{J}_{1\oplus}\mathbf{C}(\mathbf{x}_R, \mathbf{x}_i).$$

$\mathbf{A}'$ is the covariance matrix representing the uncertainty in the new location of the robot. $\mathbf{B}'$ is a row in the system variance matrix. The $i$th element is a sub-matrix — the cross-covariance of the robot's estimated location and the estimated location of the $i$th object, as given above. If the estimates of the two locations were not dependent, then that sub-matrix was, and remains $\mathbf{0}$. The newly estimated cross-covariance matrices are transposed, and written into the $R$th column of the system variance matrix, marked by $\mathbf{B}'^T$.

## 4.2 New Spatial Information

The second process which changes the map is the update that occurs when new information about the system state is incorporated. New spatial information might be given, determined by sensor measurements, or even deduced as the consequence of applying a geometrical constraint. For example, placing a box on a table reduces the degrees of freedom of the box and eliminates the uncertainties in the lost degrees of freedom (with respect to the table coordinate frame). In our example, state information is obtained as prior knowledge, or through measurement.

There are two cases which arise when adding new spatial information about objects to our map:

- I: A new object is added to the map,
- II: A (stochastic) constraint is added between objects already in the map.

We will consider each of these cases in turn.

### 4.2.1 Case I: Adding New Objects

When a new object is added to the map, a new entry must be made in the system state vector to describe the object's *world* location. A new row and column are also added to the system variance matrix to describe the uncertainty in the object's estimated location, and the inter-dependencies of this estimate with estimated locations of other objects. The expanded system is:

$$\hat{\mathbf{x}}^{(+)} = \left[ \begin{array}{c} \hat{\mathbf{x}}^{(-)} \\ \hline \hat{\mathbf{x}}_{n+1} \end{array} \right], \quad \mathbf{C}(\mathbf{x}^{(+)}) = \left[ \begin{array}{c|c} \mathbf{C}(\mathbf{x}^{(-)}) & \mathbf{B}^T \\ \hline \mathbf{B} & \mathbf{A} \end{array} \right],$$

where $\hat{\mathbf{x}}_{n+1}$, $\mathbf{A}$, and $\mathbf{B}$ will be defined below.

We divide Case I into two sub-cases: I-a, the estimate of the new object's location is *independent* of the estimates of other object locations described in the map; or I-b, it is *dependent* on them.

Case I-a occurs when the estimated location of the object is given directly in world coordinates — i.e., $\hat{\mathbf{x}}_{new}$ and $\mathbf{C}(\mathbf{x}_{new})$ — perhaps as prior information. Since the estimate is independent of other location estimates:

$$\begin{aligned} \mathbf{x}_{n+1} &= \mathbf{x}_{new}, \\ \hat{\mathbf{x}}_{n+1} &= \hat{\mathbf{x}}_{new}, \\ \mathbf{A} = \mathbf{C}(\mathbf{x}_{n+1}) &= \mathbf{C}(\mathbf{x}_{new}), \\ \mathbf{B}_i = \mathbf{C}(\mathbf{x}_{n+1}, \mathbf{x}_i) &= \mathbf{C}(\mathbf{x}_{new}, \mathbf{x}_i) = \mathbf{0}. \end{aligned} \tag{13}$$

where $\mathbf{A}$ is a covariance matrix, and $\mathbf{B}$ is a row of cross-covariance matrices, as before. $\mathbf{B}$ is identically $\mathbf{0}$, since the new estimate is independent of the previous estimates, by definition.

Case I-b occurs when the *world* location of the new object is determined as a function, $\mathbf{g}$, of its spatial relation, $\mathbf{z}$, to other object locations estimated in the map. The relation might be measured or given as prior information. For example, the robot measures the location of a new object relative to itself. Clearly, the uncertainty in the object's *world* location is correlated with the uncertainty in the robot's (world) location. For Case I-b:

$$\begin{aligned} \mathbf{x}_{n+1} &= \mathbf{g}(\mathbf{x}, \mathbf{z}), \\ \hat{\mathbf{x}}_{n+1} &= \mathbf{g}(\hat{\mathbf{x}}, \hat{\mathbf{z}}), \\ \mathbf{A} = \mathbf{C}(\mathbf{x}_{n+1}) &= \mathbf{G_x}\mathbf{C}(\mathbf{x})\mathbf{G_x}^T + \mathbf{G_y}\mathbf{C}(\mathbf{z})\mathbf{G_y}, \\ \mathbf{B}_i &= \mathbf{C}(\mathbf{x}_{n+1}, \mathbf{x}_i), \\ \mathbf{B} &= \mathbf{G_x}\mathbf{C}(\mathbf{x}). \end{aligned} \tag{14}$$

We see that Case I-a is the special case of Case I-b, where estimates of the world locations of new objects are independent of the old state estimates and are given exactly by the measured information. That is, when:

$$\mathbf{g}(\mathbf{x}, \mathbf{z}) = \mathbf{z}.$$

### 4.2.2 Case II: Adding Constraints

When new information is obtained relating objects *already in the map*, the system state vector and variance matrix do not increase in size; i.e., no new elements are introduced. However, the old elements are *constrained* by the new relation, and their values will be changed. Constraints can arise in a number of ways:

- A robot measures the relationship of a *known* landmark to itself (i.e., estimates of the world locations of robot and landmark already exist).
- A geometric relationship, such as colinearity, coplanarity, etc., is given for some set of the object location variables.

In the first example the constraint is noisy (because of an imperfect measurement). In the second example, the constraint could be absolute, but could also be given with a tolerance. The two cases are mathematically similar, in that they have to do with uncertain relationships on a number of variables — either measured, or hypothesized. A "rectangularity" constraint is discussed later in the example.

When a constraint is introduced, there are two estimates of the geometric relationship in question — our current best estimate of the relation, which can be extracted from the map, and the new information. The two estimates can be compared (in the same reference frame), and together should allow some improved estimate to be formed (as by averaging, for instance).

For each sensor, we have a *sensor model* that describes how the sensor maps the spatial variables in the state vector into sensor variables. Generally, the measurement, $\mathbf{z}$, is described as a function, $\mathbf{h}$, of the state vector, corrupted by mean-zero, additive noise $\mathbf{v}$. The covariance of the noise, $\mathbf{C}(\mathbf{v})$, is given as part of the model.

$$\mathbf{z} = \mathbf{h}(\mathbf{x}) + \mathbf{v}. \tag{15}$$

The *conditional* sensor value, given the state, and the *conditional covariance* are easily estimated from (15) as:

$$\hat{\mathbf{z}} \approx \mathbf{h}(\hat{\mathbf{x}}).$$

$$\mathbf{C}(\mathbf{z}) \approx \mathbf{H_x}\mathbf{C}(\mathbf{x})\mathbf{H_x}^T + \mathbf{C}(\mathbf{v}),$$

where:

$$\mathbf{H_x} \triangleq \frac{\partial \mathbf{h}_k(\mathbf{x})}{\partial \mathbf{x}} \left(\hat{\mathbf{x}}_k^{(-)}\right)$$

The formulae describe what values we *expect* from the sensor under the circumstances, and the likely variation; it is our current best estimate of the relationship to be measured. The actual sensor values returned are usually assumed to be conditionally independent of the state, meaning that the noise is assumed to be independent in each measurement, even when measuring the same relation with the same sensor. The actual sensor values, corrupted by the noise, are the second estimate of the relationship.

For simplicity, in our example we assume that the sensor measures the relative location of the observed object in Cartesian coordinates. Thus the sensor function becomes the tail-to-tail relation of the location of the sensor and the sensed object, described in Section 3.2.3.

(Formally, the sensor function is a function of all the variables in the state vector, but the unused variables are not shown below):

$$\mathbf{z} = \mathbf{x}_{ij} = \ominus\mathbf{x}_i \oplus \mathbf{x}_j.$$

$$\hat{\mathbf{z}} = \hat{\mathbf{x}}_{ij} = \ominus\hat{\mathbf{x}}_i \oplus \hat{\mathbf{x}}_j.$$

$$\mathbf{C}(\mathbf{z}) = {}_{\ominus}\mathbf{J}_{\oplus} \begin{bmatrix} \mathbf{C}(\mathbf{x}_i) & \mathbf{C}(\mathbf{x}_i, \mathbf{x}_j) \\ \mathbf{C}(\mathbf{x}_j, \mathbf{x}_i) & \mathbf{C}(\mathbf{x}_j) \end{bmatrix} {}_{\ominus}\mathbf{J}_{\oplus}^T + \mathbf{C}(\mathbf{v}).$$

Given the sensor model, the conditional estimates of the sensor values and their uncertainties, and an actual sensor measurement, we can update the state estimate using the Kalman Filter equations [Gelb, 1984] given below, and described in the next section:

$$\begin{aligned} \hat{\mathbf{x}}_k^{(+)} &= \hat{\mathbf{x}}_k^{(-)} + \mathbf{K}_k \left[\mathbf{z}_k - \mathbf{h}_k(\hat{\mathbf{x}}_k^{(-)})\right], \\ \mathbf{C}(\mathbf{x}_k^{(+)}) &= \mathbf{C}(\mathbf{x}_k^{(-)}) - \mathbf{K}_k \mathbf{H}_{\mathbf{x}} \mathbf{C}(\mathbf{x}_k^{(-)}), \\ \mathbf{K}_k &= \mathbf{C}(\mathbf{x}_k^{(-)}) \mathbf{H}_{\mathbf{x}}^T \left[\mathbf{H}_{\mathbf{x}} \mathbf{C}(\mathbf{x}_k^{(-)}) \mathbf{H}_{\mathbf{x}}^T + \mathbf{C}(\mathbf{v})_k\right]^{-1}. \end{aligned} \tag{16}$$

### 4.2.3 Kalman Filter

The updated estimate is a weighted average of the two estimates, where the weighting factor (computed in the weight matrix $\mathbf{K}$) is proportional to the prior covariance in the state estimate, and inversely proportional to the conditional covariance of the measurement. Thus, if the measurement covariance is large, compared to the state covariance, then $\mathbf{K} \rightarrow \mathbf{0}$, and the measurement has little impact in revising the state estimate. Conversely, when the prior state covariance is large compared to the noise covariance, then $\mathbf{K} \rightarrow \mathbf{I}$, and nearly the entire difference between the measurement and its expected value is used in updating the state.

The Kalman Filter generally contains a system dynamics model defined less generally than presented in (10); in the standard filter equations the process noise is additive:

$$\mathbf{x}_k^{(-)} = \mathbf{f}\left(\mathbf{x}_{k-1}^{(+)}, \mathbf{u}_{k-1}\right) + \mathbf{w}_{k-1} \tag{17}$$

in that case $\mathbf{F}_{\mathbf{y}}$ of (10) is the identity matrix, and the estimated mean and covariance take the form:

$$\hat{\mathbf{x}}_k^{(-)} \approx \mathbf{f}\left(\hat{\mathbf{x}}_{k-1}^{(+)}, \mathbf{u}_{k-1}\right), \tag{18}$$

$$\mathbf{C}(\mathbf{x}_k^{(-)}) \approx \mathbf{F}_{\mathbf{x}} \mathbf{C}(\mathbf{x}_{k-1}^{(+)}) \mathbf{F}_{\mathbf{x}}^T + \mathbf{C}(\mathbf{w}_{k-1}).$$

If the functions $\mathbf{f}$ in (17) and $\mathbf{h}$ in (15) are *linear* in the state vector variables, then the partial derivative matrices $\mathbf{F}$ and $\mathbf{H}$ are simply constants, and the update formulae (16) with (17), (15), and (18), represent the Kalman Filter [Gelb, 1984].

If, in addition, the noise variables are drawn from normal distributions, then the Kalman Filter produces the *optimal minimum-variance Bayesian estimate*, which is equal to the mean of the *a posteriori conditional density function* of $\mathbf{x}$, given the prior statistics of $\mathbf{x}$, and the statistics of the measurement $\mathbf{z}$. No non-linear estimator can produce estimates with smaller mean-square errors. If the noise does not have a normal distribution, then the Kalman Filter is not optimal, but produces the optimal *linear* estimate.

If the functions $\mathbf{f}$ and $\mathbf{h}$ are *non-linear* in the state variables, then $\mathbf{F}$ and $\mathbf{H}$ will have to be evaluated (they are not constant matrices). The given formulae then represent the Extended Kalman Filter, a sub-optimal non-linear estimator. It is one of the most widely used non-linear estimators because of its similarity to the optimal linear filter, its simplicity of implementation, and its ability to provide accurate estimates in practice. The error in the estimation due to the non-linearities in $\mathbf{h}$ can be greatly reduced by iteration, using the Iterated Extended Kalman Filter equations [Gelb, 1984]:

$$\hat{\mathbf{x}}_{k,i+1}^{(+)} = \hat{\mathbf{x}}_k^{(-)} + \mathbf{K}_{k,i}\left[\mathbf{z}_k - \left(\mathbf{h}_k(\hat{\mathbf{x}}_{k,i}^{(+)}) + \mathbf{H}_{\mathbf{x}}(\hat{\mathbf{x}}_k^{(-)} - \hat{\mathbf{x}}_{k,i}^{(+)})\right)\right],$$

$$\mathbf{C}(\mathbf{x}_{k,i+1}^{(+)}) = \mathbf{C}(\mathbf{x}_k^{(-)}) - \mathbf{K}_{k,i}\mathbf{H}_{\mathbf{x}}\mathbf{C}(\mathbf{x}_k^{(-)}),$$

$$\mathbf{K}_{k,i} = \mathbf{C}(\mathbf{x}_k^{(-)})\mathbf{H}_{\mathbf{x}}^T\left[\mathbf{H}_{\mathbf{x}}\mathbf{C}(\mathbf{x}_k^{(-)})\mathbf{H}_{\mathbf{x}}^T + \mathbf{C}(\mathbf{v}_k)\right]^{-1},$$

where:

$$\mathbf{H}_{\mathbf{x}} \triangleq \frac{\partial \mathbf{h}_k(\mathbf{x})}{\partial \mathbf{x}}\left(\hat{\mathbf{x}}_{k,i}^{(-)}\right)$$

$$\hat{\mathbf{x}}_{k,0}^{(+)} \triangleq \hat{\mathbf{x}}_k^{(-)}.$$

Note that the original measurement value, $\mathbf{z}$, and the prior estimates of the mean and covariance of the state, are used in each step of the iteration. The $i$th estimate of the state is used to evaluate the weight matrix, $\mathbf{K}$, and is the argument to the non-linear sensor function, $\mathbf{h}$. Iteration can be carried out until there is little further improvement in the estimate. The final estimate of the covariance need only be computed at the end of iteration, rather than at each step, since the intermediate system covariance estimates are not used.

# 5 Developed Example

The methods developed in this paper will now be applied to the mobile robot example in detail. We choose the world reference frame to be the initial location of the robot, without loss of generality. The robot's initial location with respect to the world frame is then the identity relationship (of the compounding operation), with no uncertainty.

$$\hat{\mathbf{x}} = [\hat{\mathbf{x}}_R] = [\mathbf{0}],$$

$$\mathbf{C}(\mathbf{x}) = [\mathbf{C}(\mathbf{x}_R)] = [\mathbf{0}].$$

Note, that the normal distribution corresponding to this covariance matrix (from (4)) is singular, but the limiting case as the covariance goes to zero is a dirac delta function centered on the mean estimate. This agrees with the intuitive interpretation of zero covariance implying no uncertainty.

Step 1: When the robot senses object #1, the new information must be added into the map. Normally, adding new information relative to the robot's position would fall under case I-b, but since the robot's frame is the same as the world frame, it falls under case I-a. The sensor returns the mean location and variance of object #1 ($\hat{\mathbf{z}}_1$ and $\mathbf{C}(\mathbf{z}_1)$). The new system state vector and variance matrix are:

$$\hat{\mathbf{x}} = \begin{bmatrix} \hat{\mathbf{x}}_R \\ \hat{\mathbf{x}}_1 \end{bmatrix} = \begin{bmatrix} \mathbf{0} \\ \hat{\mathbf{z}}_1 \end{bmatrix},$$

$$\mathbf{C}(\mathbf{x}) = \begin{bmatrix} \mathbf{C}(\mathbf{x}_R) & \mathbf{C}(\mathbf{x}_R, \mathbf{x}_1) \\ \mathbf{C}(\mathbf{x}_1, \mathbf{x}_R) & \mathbf{C}(\mathbf{x}_1) \end{bmatrix} = \begin{bmatrix} \mathbf{0} & \mathbf{0} \\ \mathbf{0} & \mathbf{C}(\mathbf{z}_1) \end{bmatrix}.$$

where $\mathbf{x}_1$ is the location of object #1 with respect to the world frame.

Step 2: The robot moves from its current location to a new location, where the relative motion is given by $\mathbf{y}_R$. Since this motion is also from the world frame, it is a special case of the dynamics extrapolation.

$$\hat{\mathbf{x}} = \begin{bmatrix} \hat{\mathbf{x}}_R \\ \hat{\mathbf{x}}_1 \end{bmatrix} = \begin{bmatrix} \hat{\mathbf{y}}_R \\ \hat{\mathbf{z}}_1 \end{bmatrix},$$

$$\mathbf{C}(\mathbf{x}) = \begin{bmatrix} \mathbf{C}(\mathbf{x}_R) & \mathbf{C}(\mathbf{x}_R, \mathbf{x}_1) \\ \mathbf{C}(\mathbf{x}_1, \mathbf{x}_R) & \mathbf{C}(\mathbf{x}_1) \end{bmatrix} = \begin{bmatrix} \mathbf{C}(\mathbf{y}_R) & \mathbf{0} \\ \mathbf{0} & \mathbf{C}(\mathbf{z}_1) \end{bmatrix}.$$

We can now transform the information in our map from the world frame to the robot's new frame to see how the world looks from the robot's point of view:

$$\begin{aligned} \hat{\mathbf{x}}_{RW} &= \ominus \hat{\mathbf{x}}_R, \\ \mathbf{C}(\mathbf{x}_{RW}) &\approx \mathbf{J}_\ominus \mathbf{C}(\mathbf{x}_R) \mathbf{J}_\ominus^T, \\ \hat{\mathbf{x}}_{R1} &= \ominus \hat{\mathbf{x}}_R \oplus \hat{\mathbf{x}}_1, \\ \mathbf{C}(\mathbf{x}_{R1}) &\approx \mathbf{J}_{1\oplus} \mathbf{J}_\ominus \mathbf{C}(\mathbf{x}_R) \mathbf{J}_\ominus^T \mathbf{J}_{1\oplus}^T + \mathbf{J}_{2\oplus} \mathbf{C}(\mathbf{x}_1) \mathbf{J}_{2\oplus}^T. \end{aligned}$$

Step 3: The robot now senses an object from its new location. The new measurement, $\mathbf{z}_2$, is of course, relative to the robot's location, $\mathbf{x}_R$.

$$\begin{aligned} \hat{\mathbf{x}} &= \begin{bmatrix} \hat{\mathbf{x}}_R \\ \hat{\mathbf{x}}_1 \\ \hat{\mathbf{x}}_2 \end{bmatrix} = \begin{bmatrix} \hat{\mathbf{y}}_R \\ \hat{\mathbf{z}}_1 \\ \hat{\mathbf{y}}_R \oplus \hat{\mathbf{z}}_2 \end{bmatrix}, \\ \mathbf{C}(\mathbf{x}) &= \begin{bmatrix} \mathbf{C}(\mathbf{x}_R) & \mathbf{C}(\mathbf{x}_R, \mathbf{x}_1) & \mathbf{C}(\mathbf{x}_R, \mathbf{x}_2) \\ \mathbf{C}(\mathbf{x}_1, \mathbf{x}_R) & \mathbf{C}(\mathbf{x}_1) & \mathbf{C}(\mathbf{x}_1, \mathbf{x}_2) \\ \mathbf{C}(\mathbf{x}_2, \mathbf{x}_R) & \mathbf{C}(\mathbf{x}_2, \mathbf{x}_1) & \mathbf{C}(\mathbf{x}_2) \end{bmatrix} \\ &= \begin{bmatrix} \mathbf{C}(\mathbf{y}_R) & \mathbf{0} & \mathbf{C}(\mathbf{y}_R)\mathbf{J}_{1\oplus}^T \\ \mathbf{0} & \mathbf{C}(\mathbf{z}_1) & \mathbf{0} \\ \mathbf{J}_{1\oplus}\mathbf{C}(\mathbf{y}_R) & \mathbf{0} & \mathbf{C}(\mathbf{x}_2) \end{bmatrix}. \end{aligned}$$

where:

$$\mathbf{C}(\mathbf{x}_2) = \mathbf{J}_{1\oplus} \mathbf{C}(\mathbf{y}_R) \mathbf{J}_{1\oplus}^T + \mathbf{J}_{2\oplus} \mathbf{C}(\mathbf{z}_2) \mathbf{J}_{2\oplus}^T.$$

Step 4: Now, the robot senses object #1 again. In practice one would probably calculate the world location of a new object, and only after comparing the new object to the old ones could the robot decide that they are likely to be the same object. For this example, however, we will assume that the sensor is able to identify the object as being object #1 and we don't need to map this new measurement into the world frame before performing the update. The symbolic expressions for the estimates of the mean and covariance of the state vector become too complex to reproduce as we have done for the previous steps. Also, if the iterated methods are being used, there is no symbolic expression for the results.

Notice that the formulae presented in this section are correct for *any* network of relationships which has the same topology as this example. This procedure can be completely automated, and is very suitable for use in off-line robot planning.

As a further example of some of the possibilities of this stochastic map method, we will present an example of a geometric constraint — four points known to be arranged in a rectangle. Figure 6 shows the estimated locations of the four points with respect to the world frame, before and after introduction of the information that they are the vertices of a rectangle. The improved estimates are overlayed on the original estimates in the "after" diagram. One way to specify the "rectangularity" of four points — $\mathbf{x}_i, \mathbf{x}_j, \mathbf{x}_k, \mathbf{x}_l$ is as follows:

$$\mathbf{h} = \begin{bmatrix} x_i - x_j + x_k - x_l \\ y_i - y_j + y_k - y_l \\ (x_i - x_j)(x_k - x_j) + (y_i - y_j)(y_k - y_j) \end{bmatrix}.$$

The first two elements of $\mathbf{h}$ are zero when opposite sides of the closed planar figure represented by the four vertices are parallel; the last element of $\mathbf{h}$ is zero when the two sides forming the upper–right corner are perpendicular.

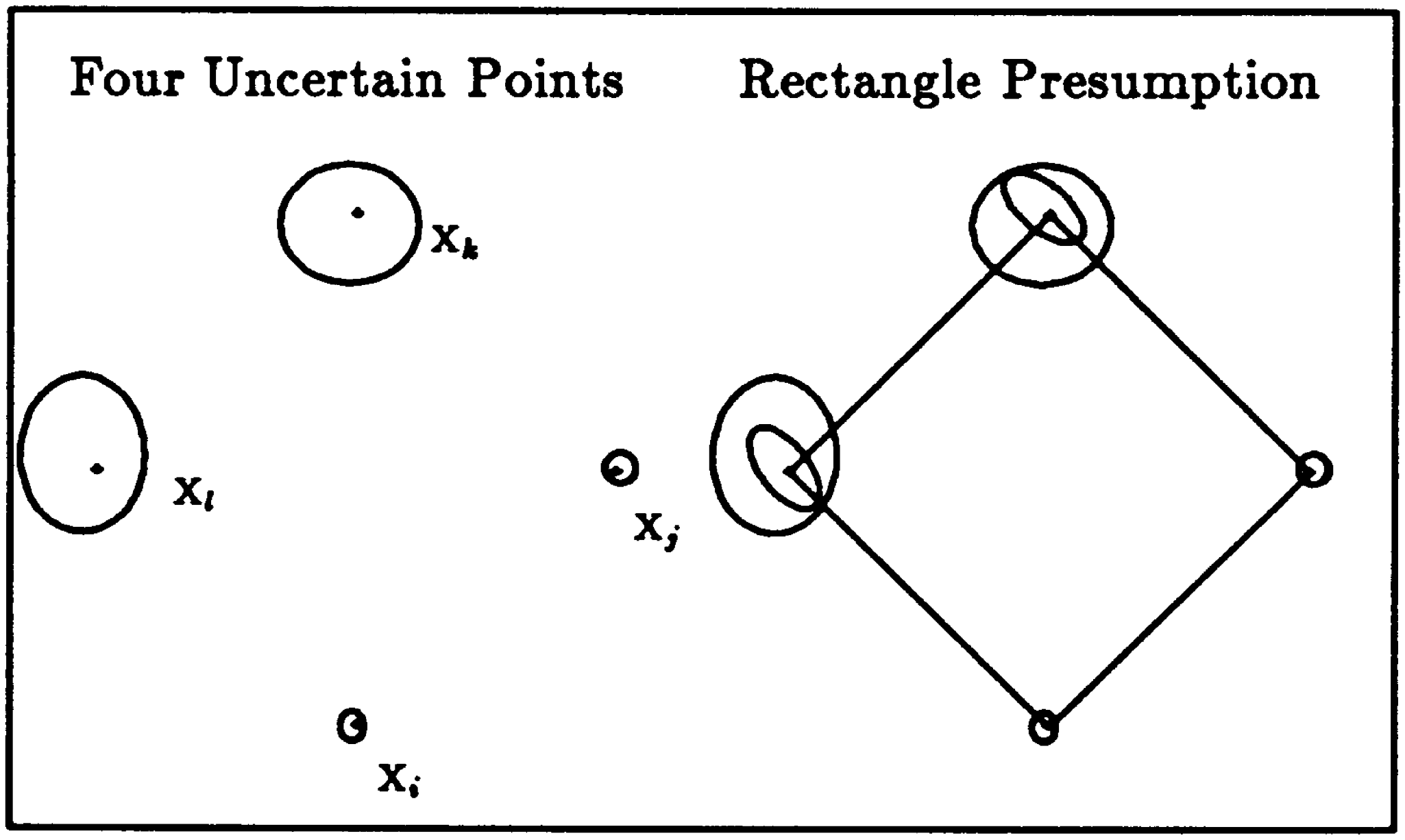

**Figure 6: Application of a Rectangular Shape Constraint**

We model the rectangle constraint similarly to a sensor, except that we hypothesize rather than measure the relationship. Just as the sensor model included measurement noise, this shape constraint could be "noisy", but here the "noise" describes random tolerances in the shape parameters, possibly given in the geometric model of the object:

$$\mathbf{z} = \mathbf{h}(\mathbf{x}) + v.$$

Given four estimated points, their nominal rectangularity ($\hat{\mathbf{z}}$) and the estimated covariance can be computed. The new information — the presumed shape — is chosen with shape parameters from a distribution with mean $\mathbf{0}$ and covariance $\mathbf{C}(\mathbf{v})$. We might as well choose the most likely a priori value, $\mathbf{0}$.

If we are going to impose the constraint that the four points are precisely in a rectangle — i.e., there is no shape uncertainty, and $\mathbf{C}(\mathbf{v}) = \mathbf{0}$ — then we can choose $\mathbf{h}$ to be *any* function which is zero only when the four points are in a rectangle. If, however, we wish to impose a *loose* rectangle constraint, we must formulate the function $\mathbf{h}$ such that $\mathbf{z}$ is a useful measure of *how* the four points fail to be rectangular.

# 6 Discussion and Conclusions

This paper presents a general theory for estimating uncertain relative spatial relationships between reference frames in a network of uncertain spatial relationships. Such networks arise, for example, in industrial robotics and navigation for mobile robots, because the system is given spatial information in the form of sensed relationships, prior constraints, relative motions, and so on. The theory presented in this paper allows the efficient estimation of these uncertain spatial relations. This theory can be used, for example, to compute *in advance* whether a proposed sequence of actions (each with known uncertainty) is likely to fail due to too much accumulated uncertainty; whether a proposed sensor observation will reduce the uncertainty to a tolerable level; whether a sensor result is so unlikely given its expected value and its prior probability of failure that it should be ignored, and so on. This paper applies state estimation theory to the problem of estimating parameters of an entire spatial configuration of objects, with the ability to transform estimates into any frame of interest.

The estimation procedure makes a number of assumptions that are normally met in practice. These assumptions are detailed in the text, but the main assumptions can be summarized as follows:

- The angular errors are "small". This requirement arises because we linearize inherently nonlinear relationships. In Monte Carlo simulations[Smith, 1985], angular errors with a standard deviation as large as $5^o$ gave estimates of the means and variances to within 1% of the correct values.
- Estimating only two moments of the probability density functions of the uncertain spatial relationships is adequate for decision making. We believe that this is the case since we will most often model a sensor observation by a mean and variance, and the relationships which result from combining many pieces of information become rapidly Gaussian, and thus are accurately modelled by only two moments.

Although the examples presented in this paper have been solely concerned with *spatial* information, there is nothing in the theory that imposes this restriction. Provided that functions are given which describe the relationships among the components to be estimated, those components could be forces, velocities, time intervals, or other quantities in robotic and non-robotic applications.

# Appendix A

In this paper we presented formulae for computing the resultant of two spatial relationships in two dimensions (three degrees of freedom). The Jacobians for the three-dimensional transformations are described below. In three dimensions, there are six degrees of freedom: translations in $x, y, z$ and three orientation angles: $\phi, \theta, \psi$. For computational reasons, orientation is often expressed as a rotation matrix composed of orthogonal column vectors (one per Cartesian axis):

$$\mathbf{R} = \left[ \begin{array}{ccc} \mathbf{n} & \mathbf{o} & \mathbf{a} \end{array} \right] = \left[ \begin{array}{ccc} n_x & o_x & a_x \\ n_y & o_y & a_y \\ n_z & o_z & a_z \end{array} \right]$$

A primitive rotation is a rotation about one of the axes, and can be represented by a primitive rotation matrix with the above form (see [Paul, 1981] for definitions). For example, $Rot(z, a)$ describes the rotation by $a$ radians about the $z$ axis. Primitive rotation matrices can be multiplied together to produce a rotation matrix describing the final orientation. Orientation will be represented by rotation matrices in the following. There are two common interpretations of the orientation angles—Euler angles and roll, pitch, and yaw.

## Relationships Using Euler Angles

Euler angles are defined by:

$$Euler(\phi, \theta, \psi) = Rot(z, \phi) Rot(y', \theta) Rot(z'', \psi) =$$

$$\left[ \begin{array}{ccc} \cos\phi\cos\theta\cos\psi - \sin\phi\sin\psi & -\cos\phi\cos\theta\sin\psi - \sin\phi\cos\psi & \cos\phi\sin\theta \\ \sin\phi\cos\theta\cos\psi + \cos\phi\sin\psi & -\sin\phi\cos\theta\sin\psi + \cos\phi\cos\psi & \sin\phi\sin\theta \\ -\sin\theta\cos\psi & \sin\theta\sin\psi & \cos\theta \end{array} \right].$$

The head to tail relationship, $\mathbf{x}_3 = \mathbf{x}_1 \oplus \mathbf{x}_2$, is then given by:

$$\mathbf{x}_3 = \left[ \begin{array}{c} x_3 \\ y_3 \\ z_3 \\ \phi_3 \\ \theta_3 \\ \psi_3 \end{array} \right] = \left[ \begin{array}{c} \mathbf{T}_E \\ \mathbf{A}_E \end{array} \right]$$

where $\mathbf{T}_E$ and $\mathbf{A}_E$ are

$$\mathbf{T}_E = \mathbf{R}_1 \left[ \begin{array}{c} x_2 \\ y_2 \\ z_2 \end{array} \right] + \left[ \begin{array}{c} x_1 \\ y_1 \\ z_1 \end{array} \right],$$

$$\mathbf{A}_E = \left[ \begin{array}{c} atan2(a_{y_3}, a_{x_3}) \\ atan2(a_{x_3}\cos\phi_3 + a_{y_3}\sin\phi_3, a_{z_3}) \\ atan2(-n_{x_3}\sin\phi_3 + n_{y_3}\cos\phi_3, -o_{x_3}\sin\phi_3 + o_{y_3}\cos\phi_3) \end{array} \right].$$

The matrix $\mathbf{R}_1$, representing the orientation angles of $\mathbf{x}_1$, has the same definition as the Euler rotation matrix defined above (with angles subscripted by 1). The terms $a_{x_3}$ etc. are the elements of the compound rotation matrix $\mathbf{R}_3$, whose values are defined by $\mathbf{R}_3 = \mathbf{R}_1\mathbf{R}_2$. Note that the inverse trignometric function *atan*2 is a function of two arguments, the ordinate $y$ and the abscissa $x$. This function returns the correct result when either $x$ or $y$ are zero, and gives the correct answer over the entire range of possible inputs [Paul, 1981]. Also note that the solution for $\phi_3$ is obtained first, and then used in solving for the other two angles.

The Jacobian of this relationship, $\mathbf{J}_\oplus$, is:

$$\mathbf{J}_\oplus = \frac{\partial \mathbf{x}_3}{\partial(\mathbf{x}_1, \mathbf{x}_2)} = \begin{bmatrix} \mathbf{I}_{3\times3} & \mathbf{M} & \mathbf{R}_1 & \mathbf{0}_{3\times3} \\ \mathbf{0}_{3\times3} & \mathbf{K}_1 & \mathbf{0}_{3\times3} & \mathbf{K}_2 \end{bmatrix}$$

where

$$\mathbf{M} = \begin{bmatrix} -(y_3 - y_1) & (z_3 - z_1)\cos\phi_1 & o_{x_1}x_2 - n_{x_1}y_2 \\ x_3 - x_1 & (z_3 - z_1)\sin\phi_1 & o_{y_1}x_2 - n_{y_1}y_2 \\ 0 & -x_2\cos\theta_1\cos\psi_1 + y_2\cos\theta_1\sin\psi_1 - z_2\sin\theta_1 & o_{z_1}x_2 - n_{z_1}y_2 \end{bmatrix},$$

$$\mathbf{K}_1 = \begin{bmatrix} 1 & [\cos\theta_3\sin(\phi_3 - \phi_1)]/\sin\theta_3 & [\sin\theta_2\cos(\psi_3 - \psi_2)]/\sin\theta_3 \\ 0 & \cos(\phi_3 - \phi_1) & \sin\theta_2\sin(\psi_3 - \psi_2) \\ 0 & \sin(\phi_3 - \phi_1)/\sin\theta_3 & [\sin\theta_1\cos(\phi_3 - \phi_1)]/\sin\theta_3 \end{bmatrix},$$

$$\mathbf{K}_2 = \begin{bmatrix} [\sin\theta_2\cos(\psi_3 - \psi_2)]/\sin\theta_3 & [\sin(\psi_3 - \psi_2)]/\sin\theta_3 & 0 \\ \sin\theta_2\sin(\psi_3 - \psi_2) & \cos(\psi_3 - \psi_2) & 0 \\ [\sin\theta_1\cos(\phi_3 - \phi_1)]/\sin\theta_3 & [\cos\theta_3\sin(\psi_3 - \psi_2)]/\sin\theta_3 & 1 \end{bmatrix}.$$

Note that this Jacobian (and similarly, the one for RPY angles) has been simplified by the use of final terms (e.g. $x_3$, $\psi_3$). Since the final terms are computed routinely in determining the mean relationship, they are available to evaluate the Jacobian. Examination of the elements indicates the possibility of a singularity; as the mean values of the angles approach a singular combination, the accuracy of the covariance estimates using this Jacobian will decrease. Methods for avoiding the singularity during calculations are being explored.

The inverse relation, $\mathbf{x}'$, in terms of the elements of the relationship $\mathbf{x}$, using the Euler angle definition, is:

$$\mathbf{x}' = \begin{bmatrix} x' \\ y' \\ z' \\ \phi' \\ \theta' \\ \psi' \end{bmatrix} = \begin{bmatrix} -(n_x x + n_y y + n_z z) \\ -(o_x x + o_y y + o_z z) \\ -(a_x x + a_y y + a_z z) \\ -\psi \\ -\theta \\ -\phi \end{bmatrix}$$

where $n_x$ etc. are the elements of the rotation matrix $\mathbf{R}$ associated with the angles in the given transformation to be inverted, $\mathbf{x}$. The Jacobian of the this relationship, $\mathbf{J}_\ominus$, is:

$$\mathbf{J}_\ominus = \frac{\partial \mathbf{x}'}{\partial \mathbf{x}} = \begin{bmatrix} -\mathbf{R}^T & \mathbf{N} \\ \mathbf{0}_{3\times3} & \mathbf{Q} \end{bmatrix}, \qquad \mathbf{Q} = \begin{bmatrix} 0 & 0 & -1 \\ 0 & -1 & 0 \\ -1 & 0 & 0 \end{bmatrix},$$

$$\mathbf{N} = \begin{bmatrix} n_y x - n_x y & -n_z x\cos\phi - n_z y\sin\phi + z\cos\theta\cos\psi & y' \\ o_y x - o_x y & -o_z x\cos\phi - o_z y\sin\phi - z\cos\theta\sin\psi & -x' \\ a_y x - a_x y & -a_z x\cos\phi - a_z y\sin\phi + z\sin\theta & 0 \end{bmatrix}.$$

## Relationships Using Roll, Pitch and Yaw Angles

Roll, pitch, and yaw angles are defined by:

$$RPY(\phi,\theta,\psi) = Rot(z,\phi)Rot(y',\theta)Rot(x'',\psi) =$$

$$\begin{bmatrix} \cos\phi\cos\theta & \cos\phi\sin\theta\sin\psi - \sin\phi\cos\psi & \cos\phi\sin\theta\cos\psi + \sin\phi\sin\psi \\ \sin\phi\cos\theta & \sin\phi\sin\theta\sin\psi + \cos\phi\cos\psi & \sin\phi\sin\theta\cos\psi - \cos\phi\sin\psi \\ -\sin\theta & \cos\theta\sin\psi & \cos\theta\cos\psi \end{bmatrix}$$

The head to tail relationship, $\mathbf{x}_3 = \mathbf{x}_1 \oplus \mathbf{x}_2$, is then given by:

$$\mathbf{x}_3 = \begin{bmatrix} x_3 \\ y_3 \\ z_3 \\ \phi_3 \\ \theta_3 \\ \psi_3 \end{bmatrix} = \begin{bmatrix} \mathbf{T}_{RPY} \\ \mathbf{A}_{RPY} \end{bmatrix}$$

where $\mathbf{T}_{RPY}$ and $\mathbf{A}_{RPY}$ are defined by:

$$\mathbf{T}_{RPY} = \mathbf{R}_1 \begin{bmatrix} x_2 \\ y_2 \\ z_2 \end{bmatrix} + \begin{bmatrix} x_1 \\ y_1 \\ z_1 \end{bmatrix},$$

$$\mathbf{A}_{RPY} = \begin{bmatrix} atan2(n_{y_3}, n_{x_3}) \\ atan2(-n_{z_3}, n_{x_3}\cos\phi_3 + n_{y_3}\sin\phi_3) \\ atan2(a_{x_3}\sin\phi_3 - a_{y_3}\cos\phi_3, -o_{x_3}\sin\phi_3 + o_{y_3}\cos\phi_3) \end{bmatrix}.$$

The matrix $\mathbf{R}_1$ is the rotation matrix for the RPY angles in $\mathbf{x}_1$. The Jacobian of the head-to-tail relationship is given by:

$$\mathbf{J}_\oplus = \frac{\partial \mathbf{x}_3}{\partial(\mathbf{x}_1, \mathbf{x}_2)} = \begin{bmatrix} \mathbf{I}_{3\times3} & \mathbf{M} & \mathbf{R}_1 & \mathbf{0}_{3\times3} \\ \mathbf{0}_{3\times3} & \mathbf{K}_1 & \mathbf{0}_{3\times3} & \mathbf{K}_2 \end{bmatrix}$$

where

$$\mathbf{M} = \begin{bmatrix} -(y_3 - y_1) & (z_3 - z_1)\cos(\phi_1) & a_{x_1}y_2 - o_{x_1}z_2 \\ x_3 - x_1 & (z_3 - z_1)\sin(\phi_1) & a_{y_1}y_2 - o_{y_1}z_2 \\ 0 & -x_2\cos\theta_1 - y_2\sin\theta_1\sin\psi_1 - z_2\sin\theta_1\cos\psi_1 & a_{z_1}y_2 - o_{z_1}z_2 \end{bmatrix},$$

$$\mathbf{K}_1 = \begin{bmatrix} 1 & [\sin\theta_3\sin(\phi_3 - \phi_1)]/\cos\theta_3 & [o_{x_2}\sin\psi_3 + a_{x_2}\cos\psi_3]/\cos\theta_3 \\ 0 & \cos(\phi_3 - \phi_1) & -\cos\theta_1\sin(\phi_3 - \phi_1) \\ 0 & [\sin(\phi_3 - \phi_1)]/\cos\theta_3 & [\cos\theta_1\cos(\phi_3 - \phi_1)]/\cos\theta_3 \end{bmatrix},$$

$$\mathbf{K}_2 = \begin{bmatrix} [\cos\theta_2\cos(\psi_3 - \psi_2)]/\cos\theta_3 & [\sin(\psi_3 - \psi_2)]/\cos\theta_3 & 0 \\ -\cos\theta_2\sin(\psi_3 - \psi_2) & \cos(\psi_3 - \psi_2) & 0 \\ [a_{x_1}\cos\phi_3 + a_{y_1}\sin\phi_3]/\cos\theta_3 & [\sin\theta_3\sin(\psi_3 - \psi_2)]/\cos\theta_3 & 1 \end{bmatrix}.$$

The inverse relation, $\mathbf{x}'$, in terms of the elements of $\mathbf{x}$, using the RPY angle definition, is:

$$\mathbf{x}' = \begin{bmatrix} x' \\ y' \\ z' \\ \phi' \\ \theta' \\ \psi' \end{bmatrix} = \begin{bmatrix} -(n_x x + n_y y + n_z z) \\ -(o_x x + o_y y + o_z z) \\ -(a_x x + a_y y + a_z z) \\ atan2(o_x, n_x) \\ atan2(-a_x, n_x\cos\phi' + o_x\sin\phi') \\ atan2(n_z\sin\phi' - o_z\cos\phi', -n_y\sin\phi' + o_y\cos\phi') \end{bmatrix}$$

where $n_x$ etc. are the elements of the rotation matrix, $\mathbf{R}$, for the RPY angles in $\mathbf{x}$. The Jacobian of the inverse relationship is:

$$\mathbf{J}_{\ominus} = \frac{\partial \mathbf{x}'}{\partial \mathbf{x}} = \begin{bmatrix} -\mathbf{R}^T & \mathbf{N} \\ \mathbf{0}_{3\times 3} & \mathbf{Q} \end{bmatrix}$$

where

$$\mathbf{N} = \begin{bmatrix} n_y x - n_x y & -n_z x \cos\phi - n_z y \sin\phi + z\cos\theta & 0 \\ o_y x - o_x y & -o_z x \cos\phi - o_z y \sin\phi + z \sin\theta \sin\psi & z' \\ a_y x - a_x y & -a_z x \cos\phi - a_z y \sin\phi + z \sin\theta \cos\psi & -y' \end{bmatrix},$$

$$\mathbf{Q} = \begin{bmatrix} -a_z/(1-a_{x^2}) & -a_y \cos\phi/(1-a_{x^2}) & n_x a_x/(1-a_{x^2}) \\ a_y/(1-a_{x^2})^{1/2} & -a_z \cos\phi/(1-a_{x^2})^{1/2} & o_x/(1-a_{x^2})^{1/2} \\ a_z a_x/(1-a_{x^2}) & -o_x \cos\psi/(1-a_{x^2}) & -n_x/(1-a_{x^2}) \end{bmatrix}.$$